\documentclass[12pt]{iopart}

\usepackage{iopams}  
\usepackage{graphicx}
\usepackage[dvipsnames]{xcolor}
\usepackage{enumitem}
\usepackage{array}
\usepackage{hyperref}
\hypersetup{
    colorlinks=true,
    linkcolor=blue,
    filecolor=blue,      
    urlcolor=blue,
    }

\begin{document}

\title[Omnidirectional catadioptric vision sensors]{Omnidirectional vision sensors based on catadioptric systems with discrete infrared photoreceptors for swarm robotics}

\author{Jose Fernando Contreras-Monsalvo, Victor Dossetti$^*$ and Blanca Susana Soto-Cruz}

\address{Centro de Investigaci\'on en Dispositivos Semiconductores - Instituto de Ciencias, Benem\'erita Universidad Aut\'onoma de Puebla, Av. San Claudio esq. 14 Sur, Edif. IC6, Puebla, Pue. 72570, Mexico \\
$^*$Author to whom any correspondence should be addressed.}
\ead{jose.contrerasmo@alumno.buap.mx, victor.dossetti@correo.buap.mx, vdossetti@gmail.com and blanca.soto@correo.buap.mx}

\vspace{10pt}
\begin{indented}
\item[]April 2025
\end{indented}

\begin{abstract}
In this work, we fabricated and studied two designs for omnidirectional vision sensors for swarm robotics, based on catadioptric systems consisting of a mirror with rotational symmetry, eight discrete infrared photodiodes and a single LED, in order to provide localization and navigation abilities for mobile robotic agents. We considered two arrangements for the photodiodes: one in which they point upward into the mirror, and one in which they point outward, perpendicular to the mirror. To determine which design offers a better field of view on the plane, as well as detection of distance and orientation between two agents, we developed a test rail with three degrees of freedom to experimentally and systematically measure the signal registered by the photodiodes of a given sensor (in a single readout) from the light emitted by another as functions of the distance and orientation. Afterwards, we processed and analyzed the experimental data to develop mathematical models for the mean response of a photodiode in each design. Finally, by numerically inverting the models, we compared the two designs in terms of their accuracy. Our results show that the design with the photodiodes pointing upward resolves better the distance, while the other resolves better the orientation of the emitting agent, both providing an omnidirectional field of view.
\end{abstract}

\vspace{2pc}
\noindent{\it Keywords}: computer vision, catadioptric sensors, swarm robotics

\section{Introduction}

Localization is a key factor in the implementation of navigation and motion control in mobile autonomous robotic systems. Several methods can be implemented for this purpose. For instance, the Global Positioning System (GPS) is a straightforward solution that is suitable for robots moving outdoors. Alternatively, methods based on internal sensors (odometry), map based and Simultaneous Localization and Mapping (SLAM), may be better suited for robots that operate indoors, where GPS systems are less reliable \cite{li14, ke21}. Regarding internal computer-vision systems, if the size, power, and computational capabilities of the robotic agents allow it, camera-based catadioptric sensors have been usually integrated for many years, as some of these can provide an omnidirectional field-of-view (FOV) with long-range detection \cite{spa05, tan14, bar18, che20, pat24, had24}. Accordingly, several types of mobile robots of different sizes and designs have been developed for many different applications spanning office, military, hospital, industrial, hazardous, and agricultural environments, among others \cite{luo03}.

The development of autonomous robotic systems has allowed the development of the field of \emph{swarm robotics}. Swarm robotics is a technical approach in which multiple robots collaborate to simultaneously address multiple or specific problems through local interaction and cooperation of the agents in the group \cite{bra13, kar20, dus22, sha23}. Usually, all units in the swarm share the same design, size and capabilities, which favors the development of small mobile robots with decentralized control. This is achieved by providing specific interaction rules for each robotic agent, that are continuously executed in an infinite loop, leading to complex collective behaviors \cite{dud01}. In some cases, inexpensive to manufacture units have been used for the development of large groups, sometimes reaching hundreds and even a thousand of them (see, for example, reference \cite{rub14}). Throughout the years, various Swarm Robotics Systems (SRS) have been developed, ranging from very simple self-propelled, sensorless designs used to study complex collective dynamics \cite{yu17, deb18}, to the ultra-violet (UV) sterilization of spaces to uphold cleanliness and hygiene standards \cite{dus22}, as well as the deployment of decentralized, infrastructure-independent swarms of homogeneous aerial vehicles without explicit communication in real-world scenarios \cite{pet21}.

One of the branches of application for SRS has concentrated on the study of self-organization and emergent collective phenomena, inspired by what has been observed in biological systems \cite{car03}. Such kind of phenomena include swarming in crustaceans and insects \cite{kaw23, pen24, buh06, rom25}, bird flocking \cite{eml52, cav10}, schooling and herding in fish and mammals, respectively \cite{oku86}, firefly synchronization \cite{ram11, ram18}, chemotaxis in bacteria \cite{bag09} and phototaxis in diverse organisms \cite{jek08, ran16}, to mention a few, as well as aggregation and self-assembly in some other systems \cite{lin89}. One of the most important aspects that has come to light from these studies, is the fact that these kind of behaviors usually emerge from a few local interactions among the members of the group, i.e., the individuals in a biological group use their senses (mainly vision) to interact with their nearest neighbors and, in some cases, with a few other individuals in the periphery of the group \cite{cav10}. The kind of interactions usually observed \cite{rey87} are \emph{centering}, where the individuals direct their velocity towards the center of the group in order to maintain its cohesion. A second interaction observed is \emph{velocity alignment}, where each member of the group tries to align its velocity with that of its neighbors that, as mentioned, can be nearby or far away. Another interaction observed is \emph{collision avoidance}, where the members of the group try to avoid crashes among them. There are other interactions observed in nature, where individuals interact with external fields such as chemical gradients (\emph{chemotaxis}) \cite{bag09} or light sources (\emph{phototaxis}) \cite{jek08, ran16}. A combination of some or all of these interactions can lead to complex collective behaviors as those mentioned before. This has also been corroborated by many theoretical studies \cite{vic95, dos09, gin10, dos15}.

% ===== TABLE =====
\begin{table*}[]
\caption{\label{tab-plats} Summary of specifications for some notable SRS platforms and their vision sensors characteristics.}
\scriptsize
%\footnotesize
\begin{tabular}{@{} m{1.6cm} m{0.8cm} m{2.2cm} m{1.7cm} m{3.3cm} m{2cm} m{1.6cm} }
\br

Platform & Release Year & Processor(s) & Dimensions (Weight) & Vision Sensor & Range \hspace{5mm} (Obs\-ta\-cles) & Comm. Protocol \\

\mr

S-bot \cite{nou09} & 2001 & 400 MHz custom XScale CPU & 12 $\times$ 19 cm in diam. and height  (700 g) & Camera-based catadioptric omnidirectional plus se\-ve\-ral IR proximity sensors. & \textless 70 cm (35-90 cm) & Chain \hspace{5mm} network$^a$ \\

Alice \cite{cap05} & 2002 & PIC16F877 & 1 cubic inch & 4 IR FF$^b$ receiver sensors and 2 on the back. & 60 mm, 360$^{\circ}$ w/blind spots (\textless 3 cm) & Software \hspace{3mm} interrupts \\

Jasmine \cite{szy06} & 2006 & ATMega88 and ATMega168 & $27 \times 27 \times 35$ mm & 6 IR emitter/receiver pairs. & 60 mm, 360$^{\circ}$ w/blind spots (Yes) & PCM-filter \\

MarXbot \cite{bon10} & 2009 & 533 MHz ARM 11 & 17 $\times$ 17 cm in diam. and height (1 kg) & Camera-based catadioptric omnidirectional plus a second FF camera, i.MX31 image processing unit and several proximity sensors. & \textless 110 cm (5 m between robots) & Chain \hspace{5mm} network \\

AMiR \cite{arv09a} & 2009 & ATMega168 & 6 $\times$ 7.3 $\times$ 4.7 cm & 6 IR emitter/receiver pairs. & \textgreater 30 cm, 360$^{\circ}$ w/blind spots (3-4 cm) & B-ASK$^d$ \\

Kilobot \cite{rub12} & 2012 & ATmega 328 & 33 mm in dia\-meter & 1 IR emitter and 1 IR receiver at the bottom. & 10 cm only distance (No) & CSMA/CA$^e$ \\

Rice r-one \cite{mcl13} & 2013 & 50 MHz 32-bit ARM microcontroller & 10 cm in diameter \hbox{(230 g)} & 8 IR emitter/receiver pairs plus one IR beacon and IR obstacle detection on the sides. & 160 cm, bearing and orientation w/resolution of 22.5$^{\circ}$ (Yes) & TDMA$^f$ \\

Swarmanoid \cite{dor13} & 2013 & 533 MHz i.MX31 ARM 11, DsPIC 33-based microcontrollers & (Foot-bot) 13 cm diameter and 28 cm in height & Camera-based catadioptric omnidirectional, FF camera and 24 IR emitter/receiver pairs pointing outwards, 8 pointing down, IR distance scanner rotating module. & 10 cm - 5m \hbox{({[}40,300{]} mm}, \hbox{{[}200,1500{]} mm} w/different sensors) & Wireless \\

Colias \cite{arv14} & 2014 & 2 Atmel AVR \hspace{5mm} micro\-controllers & 4 cm in dia\-meter & 6 IR long-range emitter/receiver pairs plus 3 FF short-range emitter/receiver pairs. & 0.5-2 m (15 cm) & ASK and PSK/ASK$^g$\\

Khepera IV \cite{per16} & 2015 & 800MHz ARM Cortex-A8 & 140 mm $\times$ 58 mm in diam. and height \hbox{(540 g)}  & FF camera, 8 IR emitter/receiver pairs for obstacle detection, 4 IR emitter/receiver pairs for line following, 5 ultrasonic sensors for long range object detection. & IR 2-250 mm (IR 2-250 mm and ultrasonic 25-200 cm) & Programmable \\

Mona \cite{arv19} & 2019 & ATmega328 (expandable w/Raspberry Pi Zero) & 80 mm in dia\-meter \hbox{(45 g)} & 5 IR emitter/receiver pairs at the front (expandable with camera module). & 45 mm, 180$^{\circ}$ \hbox{(\textgreater 20 mm, 180$^{\circ}$)} & Programmable \\
\mr
\end{tabular}\\
$^{a}$In a \emph{chain network}, the robots serve as landmarks or beacons themselves; 
$^{b}$Front-facing; 
$^{c}$Pulse Code Modulation; 
$^{d}$Binary amplitude shift keying (B-ASK) modulation; 
$^{e}$Carrier Sense Multiple Access with Collision Avoidance; 
$^{f}$Time-Division Multiple Access; 
$^{g}$Amplitude Shift Keying and Phase Shift Keying.
\end{table*}
\normalsize
% ================

Regarding platforms specifically designed for research in swarm robotics, some notable examples include S-bot \cite{nou09}, Alice \cite{cap05}, Jasmine \cite{szy06}, MarXbot \cite{bon10}, AMiR \cite{arv09a}, Kilobot \cite{rub12}, Rice r-one \cite{mcl13}, Swarmanoid \cite{dor13}, Colias \cite{arv14}, Khepera IV \cite{per16} and Mona \cite{arv19}. In Table \ref{tab-plats} we present a summary of their specifications and the characteristics of their vision sensor, as well as the communication protocols used in each of them. Among these platforms, one can clearly observe two main approaches in their design philosophy. On the one hand, there is what could be considered the \emph{state-of-the-art} approach, with very complex modular robotic systems, endowed with powerful processors (typically ARM cores), actuators, power management systems and sophisticated camera-based vision sensors, aided with multiple IR and even ultrasonic sensors. Looking at Table \ref{tab-plats} it is easy to point out the platforms that follow this approach, such as S-bot, MarXbot, Swarmanoid and Khepera IV. In particular, the first three employ a camera-based catadioptric omnidirectional vision sensor, in combination with multiple IR emitter/receiver pairs distributed around and at the bottom of the robots, while Khepera IV employs a combination of IR emitter/receiver pairs, ultrasonic sensors and a front-facing camera for vision purposes. These systems provide the robots with an omnidirectional FOV with a good range, obstacle detection, communications, location and navigation services, allowing them to develop complex tasks, individually and as a group. Nonetheless, this platforms tend to be very expensive and difficult to program, limiting the number of units that can be deployed. On the other hand, there are the much more simple and inexpensive platforms such as Alice, Jasmine, AMiR, Kilobot, Colias and Mona (based on Colias); Rice r-one falls in an intermediate category regarding its power and complexity. These platforms mainly rely on a few discrete IR emitter/receiver pairs for vision, navigation, location and communication (as well as obstacle detection in some cases), typically arranged in a ``flower'' pattern, with the sensors pointing outwards and placed at the edge of the robot. The amounts and types of sensors used are limited by the processor(s) driving these units, usually Atmel or PIC microcontrollers, with a finite program memory (a few tens of kilobytes) for all their computational tasks, including motion control, vision, navigation and communication. The arrangement of the IR sensors in a ``flower'' pattern, placed at the edge of the robots, usually leads to blind spots and a limited range of detection \cite{arv09b}. Kilobot is an outlier here, employing several unconventional design choices, such as reflecting infrared light off the table surface below for communication and distance sensing up to 10 cm (about 3 robot diameters) away, with some wandering involved to determine orientation, with a great deal of focus on motion principles, actuation mechanisms and the speed virtues of coarse positioning \cite{var06}. Notwithstanding their limitations, the fact that these designs are inexpensive, with low power consumption, renders them very suitable for studies in swarm robotics with the deployment of large numbers of individuals, typically addressing self-assembly, self-organization and emergent collective behavior phenomena in laboratory conditions, where external sources of light (and noise) are controlled \cite{rub14, deb18, arv09a, eva10, kor07, bod11, cal22}.

In this work, we developed and studied two designs of omnidirectional catadioptric vision sensors consisting of one IR emitter, eight discrete receivers and one mirror. Our aim is to improve on the simple flower-like designs for small inexpensive robots described above, that use a few IR sensors pointing outwards, providing such small mobile autonomous robots with a $360^{\circ}$ planar vision with long-range detection of distance and orientation, through pairwise interactions with other units and under laboratory conditions for research in swarm robotics. For this, we concentrate on the interactions between two sensors, while obstacle and ``wall'' detection is left for the inclusion of some other systems such as line following sensors. The sensors introduced in this work could also be used for communication purposes with the use of a suitable communication protocol, allowing a single IR channel to be shared by many individulals, for example, CSMA/CA \cite{rub12}. Sensors of this kind, with a small and discrete number of receivers, are related to biologically inspired visual sensors capable of performing navigation tasks with only a few ``pixels" \cite{ser18, ful11}. Throughout the rest of the paper, we provide details on the design and fabrication of the sensors, as well as the experimental measurements performed. We also describe the processing of the experimental data. Additionally, given the computational limitations that a small mobile robot for swarm robotics may have, we developed a simple, low-computational-cost mathematical framework that can be implemented with a simple microcontroller, allowing detection from the signals obtained through the IR receivers in a single readout. Finally, we provide a coarse analysis of the accuracy of the two designs regarding distance and orientation measurements, in order to point out future directions for improving the design of camera-less catadioptric sensors with discrete receivers.

\section{Experimental details and measurements}

In this section, we outline the design process for the mirrors used in the two sensors introduced in this work, their fabrication and integration with the infrared emitter and receivers, and details on how the experimental measurements were performed. 

% ======= FIGURE =======
\begin{figure*}[t]
\includegraphics[width=\textwidth] {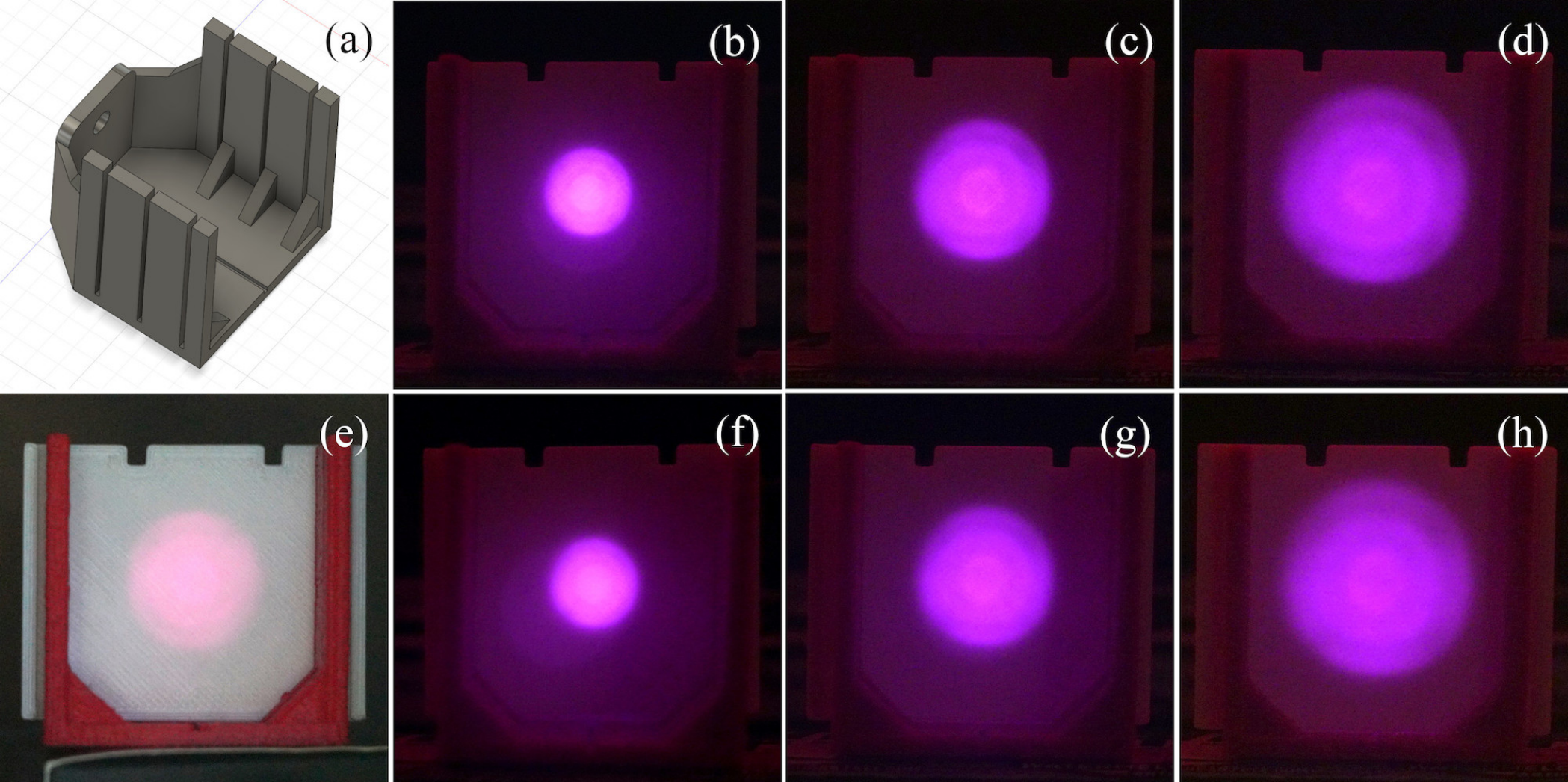}
\caption{In (a) and (e), CAD design and test jig for the characterization of the cone of light emitted by the IR LED, respectively. In (b), (c) and (d), photographs taken with an IR sensitive camera using the screen with 0.6 mm thickness at distances of 20, 30 and 40 mm from the base of the LED. In (f), (g) and (h), photographs taken using the screen with 0.8 mm thickness at distances of 20, 30 and 40 mm from the base of the LED.}
\label{fig-cones}
\end{figure*}
% =====================

\subsection{Characterization of discrete components}

As previously mentioned, each sensor consist of one IR emitter, eight discrete receivers and one mirror. As such, we selected of-the-shelf infrared emitters and receivers that are readily available and inexpensive, given the intended application of these vision sensors in SRS. Consequently, we chose as the emitter the IR333C 5 mm infrared light-emitting diode (LED) and its complementary receiver, the PD333-3B/H0/L2 5 mm silicon PIN infrared photodiode (PD), both from Everlight, with a peak emission and sensitivity wavelength of 940 nm, respectively. These devices are usually used in TV remote controls and suitable to handle 38 KHz modulated signals. According to their datasheets, the IR333C LED has a maximum DC current of 100 mA, a minimum radiant intensity of \hbox{7.8 mW/sr} (at 20 mA), a viewing angle of 40$^{\circ}$ and a stock switching frequency of 300 KHz, while the PD333-3B/H0/L2 photodiode has a viewing angle of 80$^{\circ}$ and response time of 45 ns, therefore, communication with 300 KHz modulated signals should be possible. In particular, the wide viewing angle of the PD333-3B/H0/L2 photodiode will allow us to avoid blind spots in our sensors. Nevertheless, we experimentally measured the cone of light emitted by the LED and the viewing angle of the chosen PD.

% ======= FIGURE =======
\begin{figure}[t]
\includegraphics[width=0.6\textwidth] {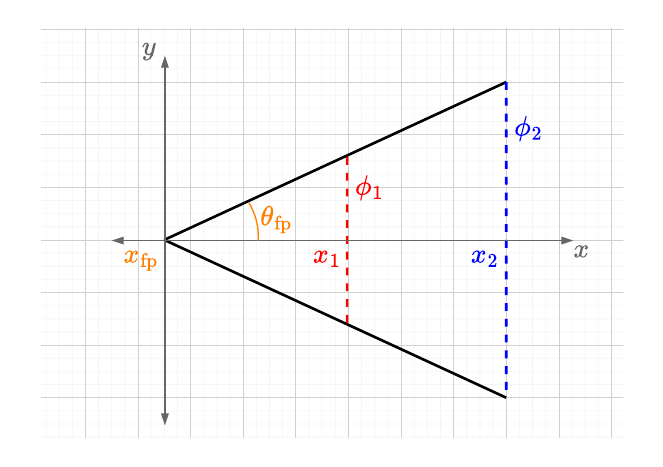}
\caption{Schematic diagram used to determine the angle of the cone $\theta_\mathrm{fp}$, as well as the apparent position $x_\mathrm{fp}$ of the focal point of the light source of the LED, from the diameter $\phi_i$ of the circle covering the illuminated area on the screen (obtained by analyzing the photographs of Fig.\ \ref{fig-cones}) at a distance $x_i$ from the front of the screen to the base of the IR LED (see the text for more details).}
\label{fig-conesDiag}
\end{figure}
% =====================

To measure the cone of light emitted by the IR LED, we first designed and 3D printed a test jig. The LED was placed in a central hole on the back wall of the jig, pointing forward, as shown in Fig.\ \ref{fig-cones}(a). The jig features three slots for positioning a white 3D printed screen, as seen in Fig.\ \ref{fig-cones}(e), at distances of 20, 30 and 40 mm measured from the LED's base. The screen includes two notches spaced 20 mm apart to calibrate the image scale for further processing. Moreover, two screen thicknesses were used, 0.6 and 0.8 mm. Figures \ref{fig-cones}(b), \ref{fig-cones}(c) and \ref{fig-cones}(d) show how the illuminated area on the 0.6 mm screen expands as the screen intersects the light cone at increasing distances. The same effect is observed in Figures \ref{fig-cones}(f), \ref{fig-cones}(g) and \ref{fig-cones}(h) for the 0.8 mm screen. Two photographs for each case were processed and analyzed using Autodesk Fusion 360. For this, the scale of each photograph was adjusted based on the distance between the screen's notches, and a circle was traced around the illuminated area to determine its diameter, $\phi_i$, as a function of the distance, $x_i$, to the LED's base; this distance included the thickness of the screen used. Afterwards, these measurements were paired as $(x_1, \phi_1)$ and $(x_2, \phi_2)$, in all possible combinations, with subscript $1$ indicating the measurement closest to the LED and subscript $2$ the furthest, to determine the angle of the cone, $\theta_\mathrm{fp}$, and the apparent position, $x_\mathrm{fp}$, of the light source, according to the schematic diagram of Fig.\ \ref{fig-conesDiag} with the equations:
\begin{eqnarray}
\theta_\mathrm{fp} = \arctan\left[\frac{\phi_2 - \phi_1}{2(x_2 - x_1)}\right], \\
x_\mathrm{fp} = x_i + \frac{\phi_i}{\tan\theta_\mathrm{fp}},
\label{cone_eqs}
\end{eqnarray}
where the subscript $i$ corresponds to the measurement 1 or 2 of a given pair. Finally, we averaged the results to obtain the mean angle of the cone to be $\theta_\mathrm{mfp} \simeq 18.4^{\circ}$, that corresponds well to the value reported in its datasheet ($20^{\circ}$), while the mean apparent position of the focal point of the LED yielded $x_\mathrm{mfp} \simeq 0$ mm, i.e., at the base of the LED.

% ======= FIGURE =======
\begin{figure}[t]
\includegraphics[width=0.6\textwidth] {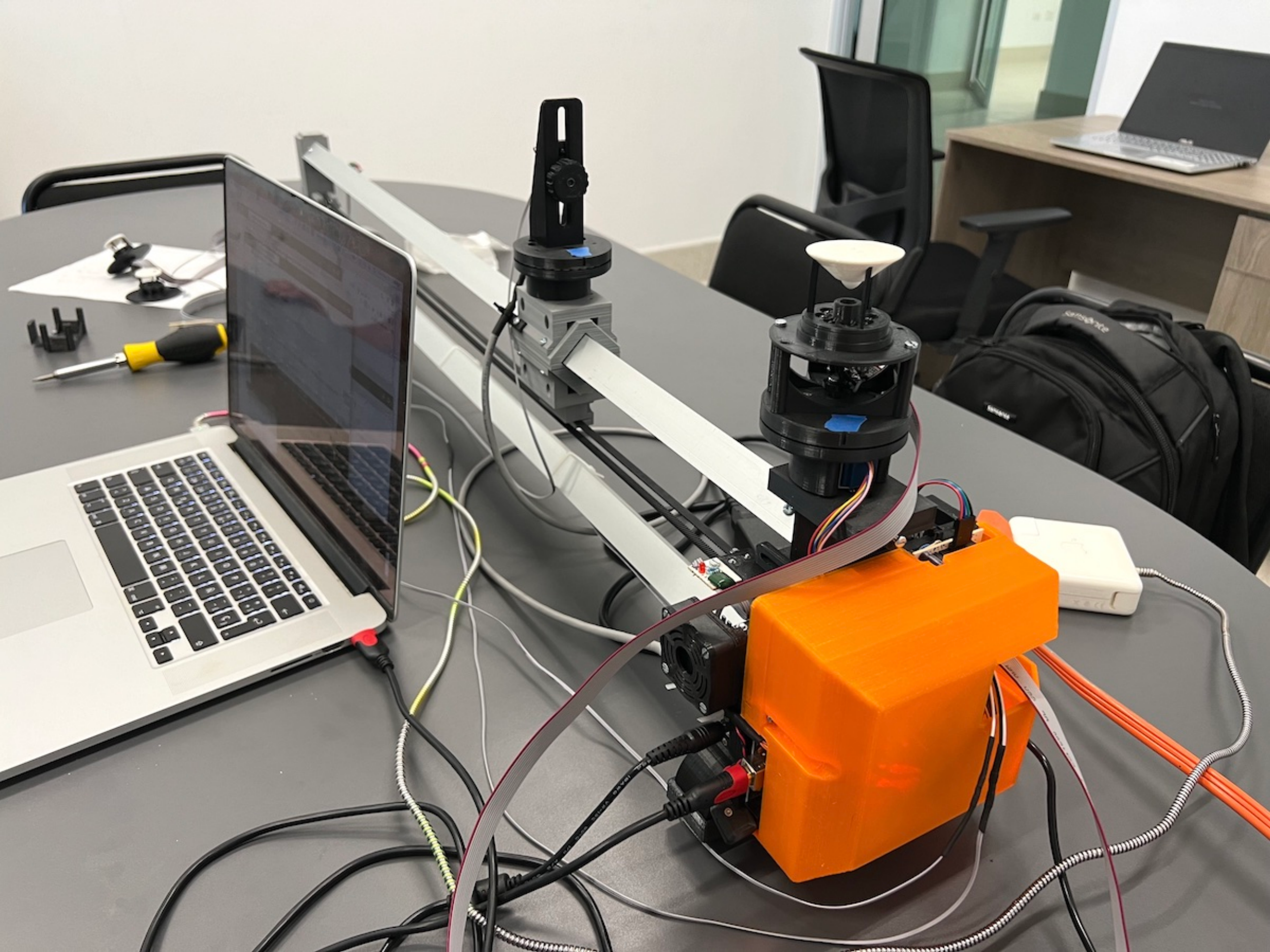}
\caption{Test stage developed in-house for all the experimental measurements performed in this work. This experimental setup includes a linear bearing that allows us to vary the distance between two bases (one mounted on the carriage of the linear bearing and one mounted on one of the supporting posts of the rail) that can rotate $360^{\circ}$, thus providing us with a total of three degrees of freedom: one longitudinal and two rotational.}
\label{fig-testSt}
\end{figure}
% =====================

On the other hand, the viewing angle of the IR photodiodes was experimentally measured using a custom designed test stage. This test stage consists of a rail built from two square aluminium tubes, supported by two 3D printed posts with stabilizing legs. Three stepper motors provide three degrees of freedom, one longitudinal with a resolution of 0.2 mm, and two rotational with a resolution of $360^{\circ}/4096$ or $0.087890625^{\circ}$ each, as shown in Fig.\ \ref{fig-testSt}. An Arduino MEGA 2560 Rev3 was used for control and data acquisition, and custom-made printed circuit boards were developed for power management, and to hold the drivers for the stepper motors and various connections. The bearings, one linear bearing mounted on the top square tube and two rotational (one mounted in one of the posts of the rail and the other on the carriage over the linear bearing), were also made of 3D printed parts, while 6 mm Airsoft BBs where used within them as rolling elements. By 3D printing different adapters, we can mount individual LEDs or PDs, as well as fully integrated sensors, to experimentally characterize their emission and reception patterns as a function of distance (up to 860 mm) and orientation (a full $360^{\circ}$).

% ======= FIGURE =======
\begin{figure}[t]
\includegraphics[width=0.6\textwidth] {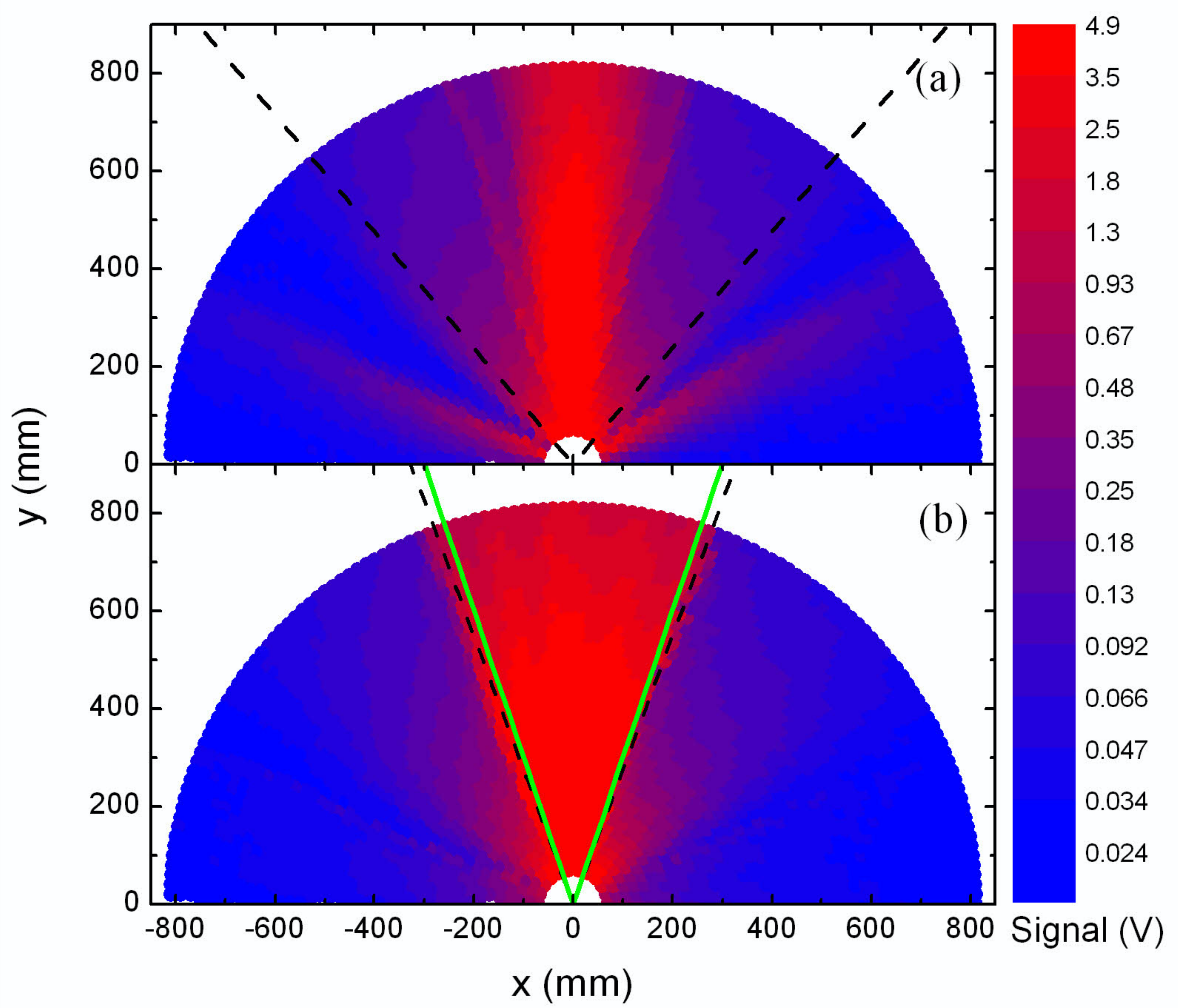}
\caption{Measured PD reception (a) and LED emission (b) patterns. A readout of the signal was obtained every 1 cm with the increasing distance, and with an angular resolution corresponding to approximately 1 cm of arc, covering $360^{\circ}$ for each case. The black dashed lines depict the corresponding viewing angles according the the datasheets, while the green lines in (b) depict the LED's cone of emission from the angle obtained with the analysis of the photographs of Fig.\ \ref{fig-cones}, using the jig with the moving screen.}
\label{fig-compPatts}
\end{figure}
% =====================

In this manner, in order to characterize the reception pattern of the PDs, we mounted one photodiode on the rotating base at the post, and an LED on the other rotating base at the carriage. This setup allowed us to measure the signal received by the PD, at different angles, as the LED, pointing directly towards the PD, was gradually moved away to record its dependency on distance. The results are shown in \hbox{Fig.\ \ref{fig-compPatts}(a)}, where red indicates a stronger signal, blue corresponds to a near-zero signal, while the black dashed lines correspond to the viewing angle given in the datasheet. The signal received by the PD was measured with a distance resolution of 1 cm and an angular resolution of approximately 1 cm of arc. Additionally, we also characterized the emission pattern of an LED with the same resolution, placing the LED on the rotating base at the post and a PD on the base at the carriage, aimed directly at the LED. These results are shown in \hbox{Fig.\ \ref{fig-compPatts}(b)}, where the green lines illustrate the emission cone determined from our study using the jig with the moving screen, while the black dashed lines correspond to the viewing angle from the datasheet. These measurements guided us in the design of the two vision sensors introduced in this work, regarding the arrangement of the discrete components, as well as the geometry and positioning of the mirrors in two ways. First, the cone of light emitted by the LED must fully cover the mirror reflecting this light. Second, the viewing angle of the photodiodes should overlap in order to avoid blind spots, wherever their disposition is in a given design. In case other devices (LEDs or PDs) with different specifications were to be employed in the design of camera-less sensors, as those of this work, their viewing angle would be of the utmost importance in their placement wthin the sensor, as well as the geometry of the mirror itself.

% ======= FIGURE =======
\begin{figure}[t]
\includegraphics[width=0.33\textwidth] {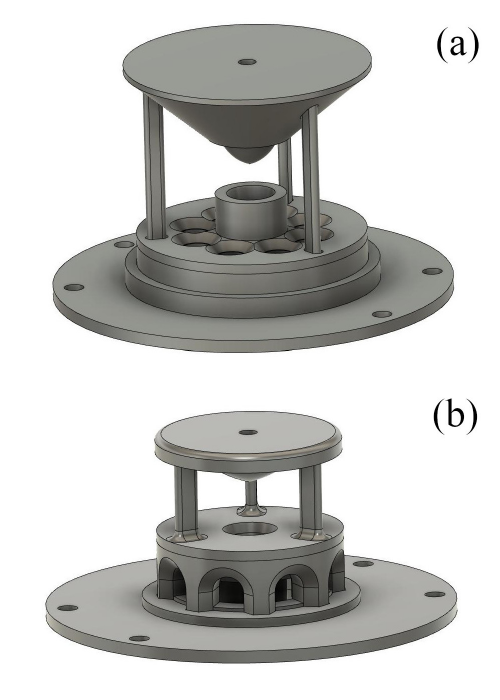}
\caption{CAD designs for the structure of the discrete components, one LED at the center, pointing upwards to the mirror, and eight PDs arranged at the base in a circular pattern and pointing upwards for the \emph{vertical} design (a), and horizontally and pointing outwards for the \emph{flower} design (b). The mirrors are supported by three thin posts on top of the discrete components.}
\label{fig-mirrorsCAD}
\end{figure}
% =====================

\subsection{Design and fabrication of the mirrors}

For the development of the vision sensors we considered two different designs. In the first, eight IR PDs are arranged in a circular formation, oriented upwards towards the mirror, as a camera would be pointed in a camera-based catadioptric vision sensor. Consequently, throughout the remainder of this article, we will refer to this as the \emph{vertical} design. In the second design, the eight PDs are arranged horizontally, aligned parallel to the mirror and directed outwards in a flower-like pattern, following the typical design of the simpler inexpensive robots described in the introducction. We will refer to this arrangement as the \emph{flower} design. In both cases, the IR LED is placed at the center of either arrangement of PDs and pointing upwards into the mirror, which itself is supported by three slender posts on top of the discrete components. Figure \ref{fig-mirrorsCAD} illustrates the CAD designs for the two configurations. 

Our aim in developing these two designs is to compare their respective resolutions in distance vs.\ orientation, as one design may prioritize distance detection at the expense of orientation resolution, while the opposite may hold true for the other design. Ultimately, our goal is to develop a complete prototype of a circular mobile robot for \emph{swarm robotics} studies, with a diameter of a soda can (approximately 66 mm) and capable of long-range (around 330 mm or five robot-diameters) vision and communication, through pairwise interactions among the agents under laboratory conditions, while border and obstacle detection would be left for other sensors to be handled (for example, line following sensor at the bottom of the robots). This, of course, lies beyond the scope of the present paper. Nonetheless, the methods introduced in this work for the development of camera-less catadioptric vision sensors with discrete receivers, may prove useful for smaller or larger prototypes with any desired range of interaction as demonstrated below. 

It is worth mentioning that the typical circuit connecting a PD to an analog-to-digital converter (ADC) includes a 10 k$\Omega$ resistor in series. During our characterization of the PDs, we observed significant variability in the peak signal generated by different PDs, along with a weakening of the LED signal once reflected by the mirror. To enhance the received signal strength and account for the PDs variable sensitivity, we added a 100 k$\Omega$ variable resistor in series with the standard 10 k$\Omega$ resistor. This allowed us to calibrate the peak signal strength uniformly across all eight PDs used in a sensor. With this calibration complete, we now focus on the design and fabrication of the mirrors, as the requirements for the two sensors differ.

% ======= FIGURE =======
\begin{figure*}[t]
\includegraphics[width=\textwidth] {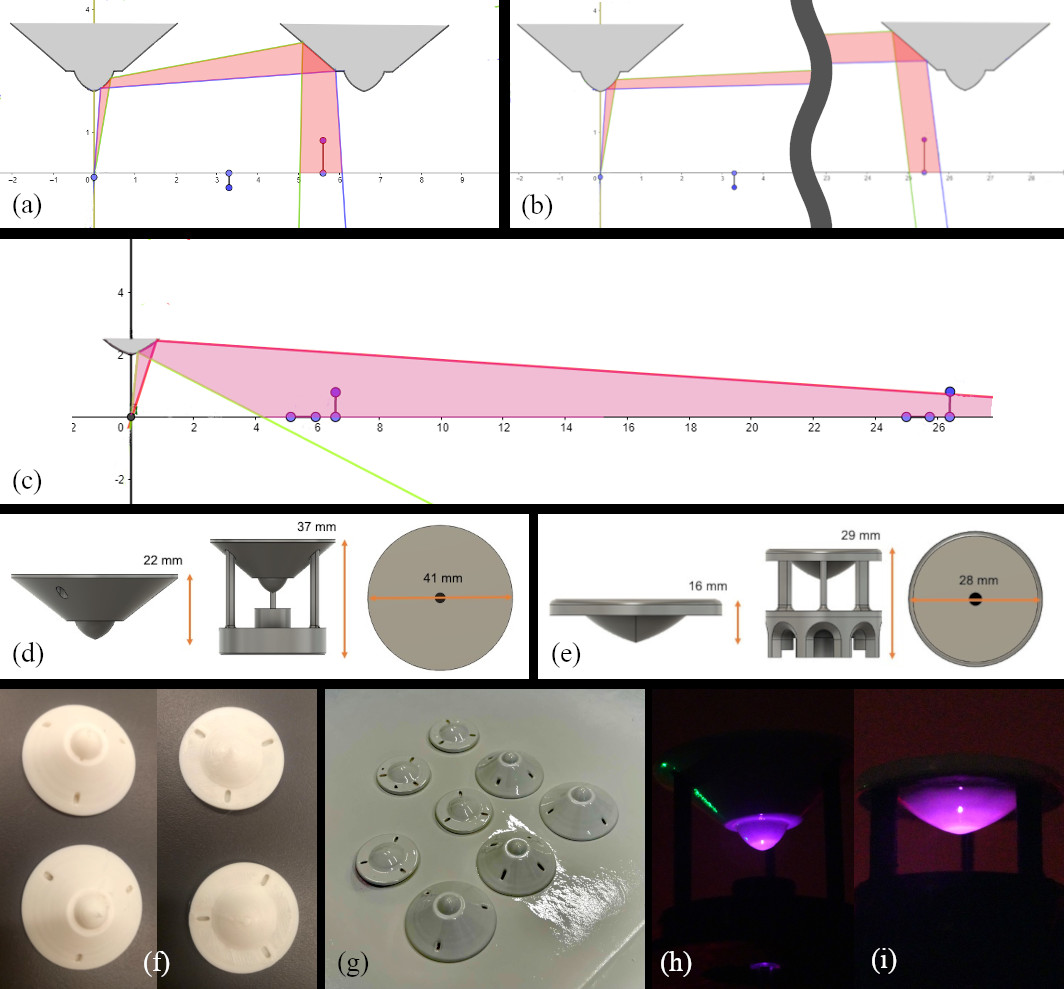}
\caption{Transverse section and geometrical optics analysis performed in GeoGebra for the \emph{vertical} design in (a) and (b) and for the \emph{flower} design in (c); see more details in the text. In (d) and (e), details of the dimensions of the mirrors, as well as their structures, for the \emph{vertical} and \emph{flower} designs, respectively. In (f), 3D printed prototypes of the mirrors. In (g), results from the smoothing process of the mirrors after polish and painting. In (h) and (i), illumination tests with the integrated sensors for the \emph{vertical} and \emph{flower} designs, respectively. These photographs were obtained with an IR sensitive camera.}
\label{fig-mirFab}
\end{figure*}
% =====================

Since the mirrors must exhibit rotational symmetry to achieve an omnidirectional FOV on the plane, only their transverse sections need to be considered. For designing the mirrors, we used GeoGebra, a software that provides tools for graphical and mathematical analysis in geometrical optics. Specifically, GeoGebra enables us to define lines, representing incoming light rays, and simulate their specular reflection from a profile defined by any function. In this way, the sensor with the \emph{vertical} design requires a two-stage mirror: one stage for the emitter and a second stage for the upward facing receivers. The shape of each stage was adjusted in GeoGebra using \emph{cubic splines} \cite{wol99}. The stage reflecting the light emitted by the LED requires a convex shape to ensure that the reflected beam covers the desired interaction range, while the second stage must be slightly concave to focus incoming light from other sensors onto the ring where the PDs are positioned. Figures \ref{fig-mirFab}(a) and \ref{fig-mirFab}(b) display the transverse section of this two-stage mirror (with the shape filled in solid grey), along with its geometrical optics analysis. In the figures, the purple vertical line with two circles under the beam (filled in translucent red) represents a PD in the receiving sensor. Figure \ref{fig-mirFab}(a) shows the analysis for a sensor placed 70 mm from the emitter, center to center, while Fig.\ \ref{fig-mirFab}(b) shows a receiving sensor at 330 mm from the emitter, center to center. It is worth mentioning that the base where each sensor is mounted has a diameter of 66 mm. By exporting the profile of the transverse section of the mirror, we generated a solid of revolution in Fusion 360, where the mirror is integrated with the structure supporting the discrete components: one LED and eight PDs. Details of the dimensions of the mirror and the support structure are presented in \hbox{Fig.\ \ref{fig-mirFab}(d)}.

On the other hand, for the \emph{flower} design, only one convex stage is required for the LED's light to reflect across the desired interaction range. In this case, the PDs are positioned horizontally around the LED and pointing outwards, thus receiving the light directly reflected from the mirror, as shown in Fig.\ \ref{fig-mirFab}(c). In this figure, purple horizontal lines with two circles represent PDs, while purple vertical lines with two circles represent LEDs. The transverse section of the mirror (shape filled in solid grey) is also shown, along with its geometrical optics analysis, with the beam of light illustrated as areas filled in translucent red. The solid of revolution and its integration with the structure for the discrete components were also modeled in Fusion 360. Details of the mirror and structural dimensions are presented in Fig.\ \ref{fig-mirFab}(e).

% ======= FIGURE =======
\begin{figure*}[t]
\includegraphics[width=\textwidth] {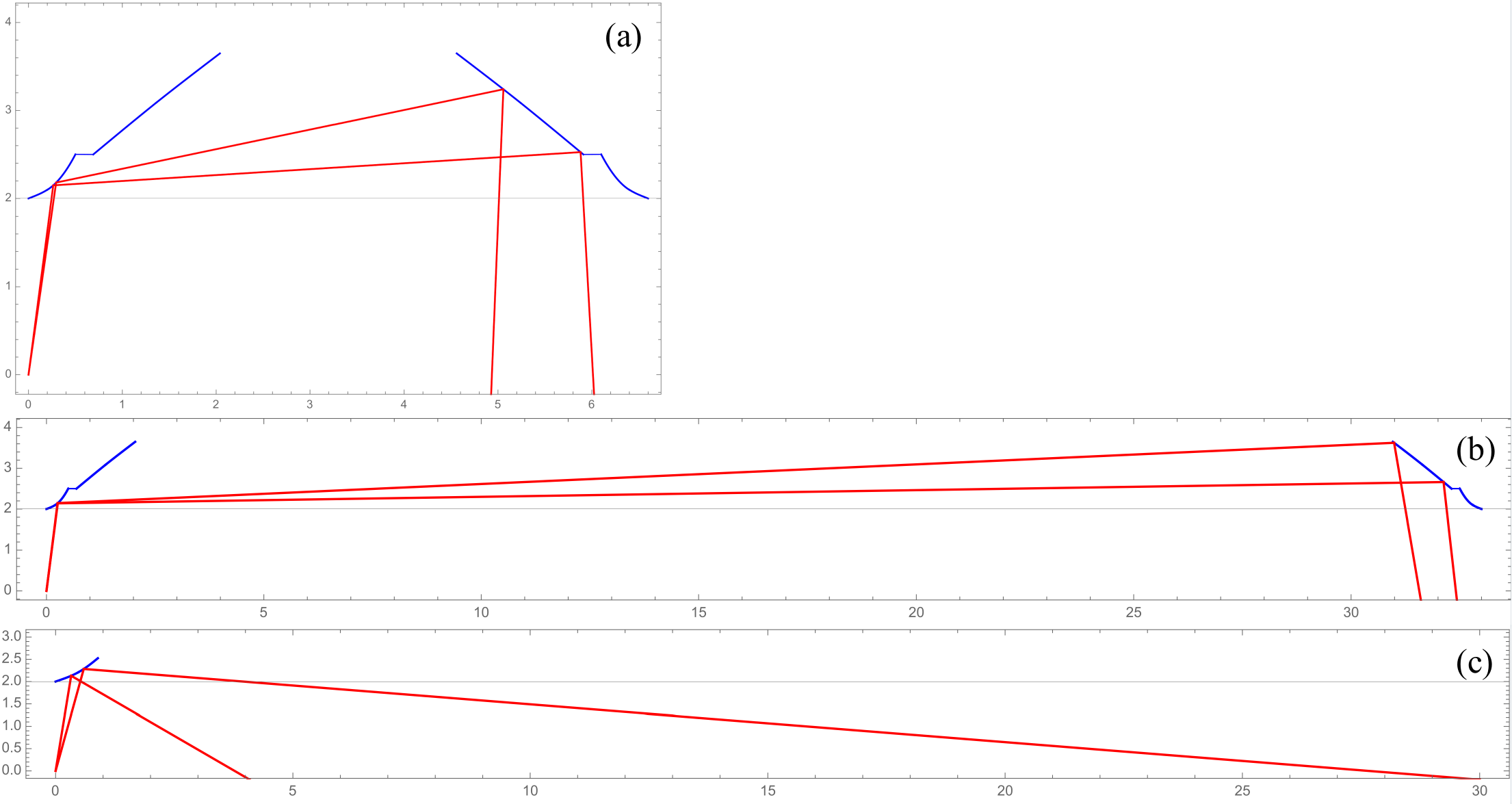}
\caption{Transverse sections of the mirrors (blue) generated using Eqs.\ (\ref{eq-vs1a}) to (\ref{eq-fb}). In this case, the geometrical optics analysis was performed in Mathematica, tracing the reflections of the different rays (red), for the \emph{vertical} design in (a) and (b), at one and five robot diameters, respectively, and for the \emph{flower} design in (c); see more details in the text.}
\label{fig-rayTrace}
\end{figure*}
% =====================

In order to validate the profiles generated by GeoGebra for the different mirror stages, we also performed ray-tracing analysis. First, we extracted the parametric functions for the splines from GeoGebra and processed them with OriginLab \emph{OriginPro} and Wolfram \emph{Mathematica} to obtain the functions for the profiles of each mirror with their corresponding stages. For the vertical design, the profile for the first stage (that reflects the light from the LED) is defined by the following piecewise function:
\begin{eqnarray}
  mp_\mathrm{v1}(x) & =  2.81723 x^3 - 0.20831 x^2 + 0.42261 x + 1.99992 \quad & x < 0.2, \label{eq-vs1a}\\
             & = 0.05936 x^3 + 2.24342 x^2 - 0.24885 x + 2.05736 \quad & x \geq 0.2, \label{eq-vs1b}
\end{eqnarray}
within the interval $x \in [0, 0.5]$, while the profile for the second stage (that focuses the incoming light on the photodiodes) corresponds to the piecewise function:
\begin{eqnarray}
  mp_\mathrm{v2}(x) & =  -0.02294 x^3 + 0.04687 x^2 + 0.8554 x + 1.89499 \quad & x < 1.36, \label{eq-vs2a}\\
             & = 0.02182 x^3 - 0.1388 x^2 + 1.1057 x + 1.78547 \quad & x \geq 1.36, \label{eq-vs2b}
\end{eqnarray}
in the interval $x \in [0.69, 2.04]$. The units for the functions $mp_\mathrm{v1}(x)$ and $mp_\mathrm{v1}(x)$, as well as $x$ correspond to centimeters. Additionally, for the profile of the single stage mirror for the flower design, we have the piecewise function:
\begin{eqnarray}
  mp_\mathrm{f}(x) & =  0.4127 x^3 - 0.07673 x^2 + 0.37849 x + 1.99975 \quad & x < 0.48, \label{eq-fa}\\
             & = -0.33283 x^3 + 1.04336 x^2 - 0.1781 x + 2.09135 \quad & x \geq 0.48, \label{eq-fb}
\end{eqnarray}
for $x \in [0, 0.89]$, with both $mp_\mathrm{f}(x)$ and $x$ measured in centimeters. Afterwards, using Mathematica and Eqs.\ (\ref{eq-vs1a}) to (\ref{eq-fb}) and their derivatives, we traced different rays and their reflections as shown in Fig.\ \ref{fig-rayTrace}. It is important to mention that this kind of analysis is mandatory to validate the profiles generated by GeoGebra, as we noticed small variations between this mathematical analysis and the results yielded by GeoGebra. Nonetheless, the profiles generated in GeoGebra were useful for our purposes in this work.

To fabricate the mirrors, the corresponding solids of revolution were 3D printed in ABS with a layer resolution of 0.1 mm, as shown in Fig.\ \ref{fig-mirFab}(f). To achieve a smooth and reflective surface, the 3D printed mirrors underwent the following steps:
\setlist[enumerate]{label={\arabic*.}}
\begin{enumerate}
\item The mirror was suspended in a closed container above a few milliliters of liquid acetone in order to receive a bath of acetone vapor for 45 minutes.
\item The mirror was sprayed with an even coat of 2-in-1 acrylic enamel with primer and left to dry for 30 minutes.
\item The mirror was mounted on an electric hand drill and received a polish with an abrasive 220 grit sanding sponge for wood.
\item Then, steps 2 and 3 were repeated at least three times until a smooth surface was achieved.
\item A final coat of paint was sprayed on the mirror and left to dry for one hour.
\end{enumerate}
The final result is shown in Fig.\ \ref{fig-mirFab}(g).  Photographs of the mirrors, integrated with the supporting structure and discrete components, were taken with an IR sensitive camera to ensure that the first stage of the mirror for the \emph{vertical} design, shown in Fig.\  \ref{fig-mirFab}(h), and the mirror for the \emph{flower} design, shown in Fig.\  \ref{fig-mirFab}(i), were fully illuminated by the LED; here we can appreciate why it was important to experimentally characterize the angle of the cone of light emitted by the LED. We must also mention that no further analyses on the quality of the surface or geometry of the mirror were performed, although the fabrication process is systematic and very reproducible, in keeping with our aim for the development of inexpensive omnidirectional vision sensors with a scalable production. However, more refined methods could produce higher quality mirrors that could improve their reflectivity and geometrical accuracy, albeit probably increasing their cost and difficulty of fabrication, as well as their production scalability. 

% ======= FIGURE =======
\begin{figure}[t]
\includegraphics[width=0.6\textwidth] {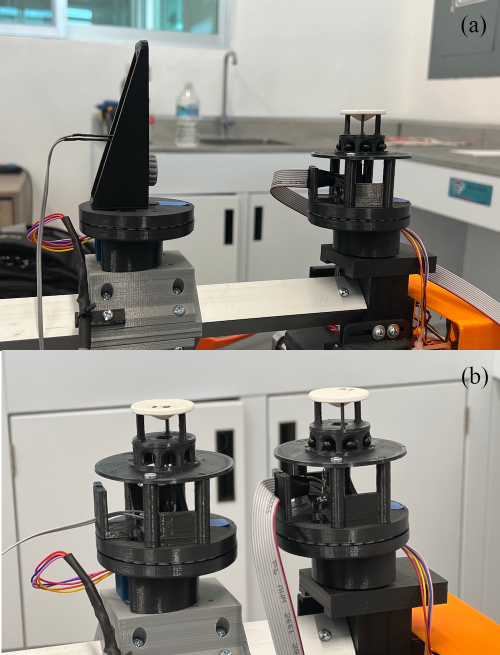}
\caption{Experimental setups for the measurement of the emission pattern of an integrated sensor (a) and for the reception pattern (b). In the first case, a sensor integrated with an LED is placed on the base at the post of the test stage, while a PD (pointing towards the sensor) is mounted on the base at the carriage. In the second case, one sensor integrated with an LED is placed on the carriage, illuminating a fully integrated sensor placed on the base at the post of the test stage.}
\label{fig-pattMeas}
\end{figure}
% =====================

% ======= FIGURE =======
\begin{figure*}[thb]
\includegraphics[width=\textwidth] {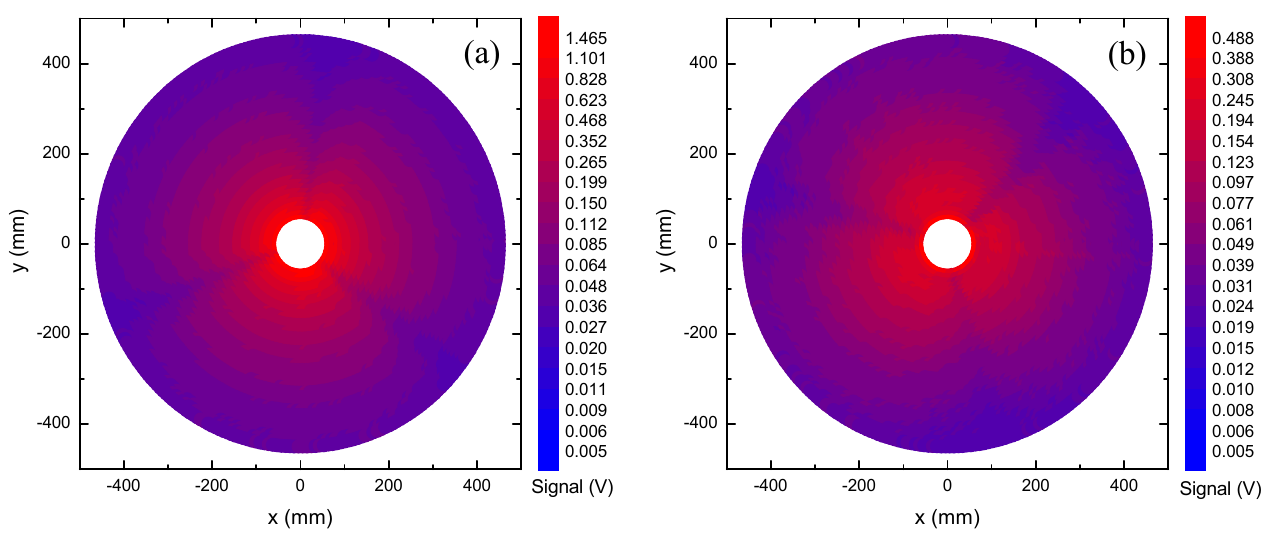}
\caption{Measured emission patterns for the \emph{vertical} (a) and \emph{flower} (b) designs, respectively. The shadows produced by the three posts that support the mirrors are apparent.}
\label{fig-emssPatt}
\end{figure*}
% =====================

\subsection{Characterization of vision sensors}

Moving forward, measurements of the emission and reception patterns were performed over the integrated vision sensors. For this purpose, we again used the test stage of Fig.\ \ref{fig-testSt}. To determine the emission pattern of a given sensor, i.e., the light from the LED reflected by the mirror, we positioned a sensor on the rotating base at the post of the test stage and a PD on the rotating base at the carriage, directed towards the sensor. The height of the PD was adjusted to target the secondary stage of the mirror for the \emph{vertical} design, and to align with the PDs for the \emph{flower} design; see Fig.\ \ref{fig-pattMeas}(a). The signal from the PD mounted at the carriage was measured during a full rotation of the sensor, with angular increments corresponding to approximately 10 mm of arc. Measurements started from a distance $d=35$ mm from the center of the sensor, increasing by 10 mm for each subsequent sweep around the full rotation of the sensor, reaching up to a distance $d=450$ mm. The results are shown in Figs.\ \ref{fig-emssPatt}(a) and \ref{fig-emssPatt}(b) for the \emph{vertical} and \emph{flower} designs, respectively. In the figures, the shadows produced by the three posts that support the mirrors are evident.

% ======= FIGURE =======
\begin{figure*}[thb]
\includegraphics[width=\textwidth] {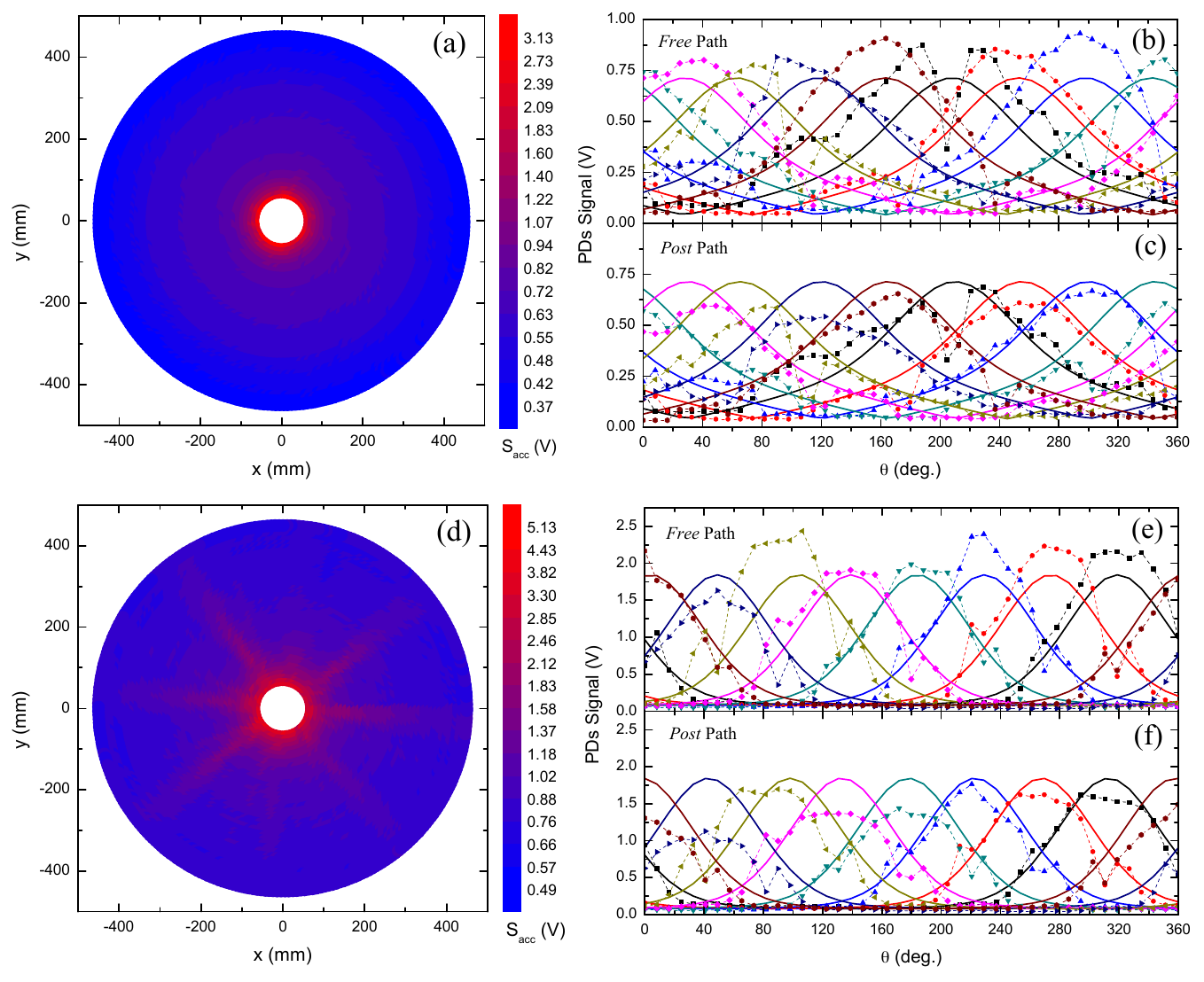}
\caption{In (a), accumulated reception pattern for the \emph{vertical} design. An example of the raw signal of each PD as a function of the orientation is shown with the dashed curves with symbols at a distance of 70 mm for the \emph{free} path in (b) and the \emph{post} path in (c) for the same design. The solid curves correspond to the model. In (d), accumulated reception pattern for the \emph{flower} design. An example of the raw signal of each PD as a function of the orientation angle $\theta$ is shown with the dashed curves with symbols for a distance of 70 mm for the \emph{free} path in (e) and the \emph{post} path in (f) for the same design. The solid curves correspond to the model developed in this work (see text for more details).}
\label{fig-recPatt}
\end{figure*}
% =====================

For the measurement of the reception patterns, two fully integrated sensors of the same type were mounted on the test stage: the one to be characterized (receiver) on the base at the post, and the other (emitter) as a source of light on the base at the carriage, as shown in Fig.\ \ref{fig-pattMeas}(b). The signal from each of the eight PDs in the receiver sensor were measured during a full rotation, with angular increments corresponding to approximately 10 mm of arc. After each full rotation, the distance between the sensors was increased by 10 mm, moving away the emitter sensor on the carriage, and another full rotation of the receiver sensor was measured under the same conditions. The range in distance covers from 70 to 450 mm, measured center-to-center between the two sensors. Two cases were considered for the measurement of the reception patterns of the sensors: one in which the LED light reflected by the mirror in the emitter sensor follows a clear path between two of the three posts supporting the mirror, here referred to as the \emph{free} path, and another where one of the posts obstructs the light path, the \emph{post} path. For this, the emitter sensor at the carriage was adjusted depending on the path of light chosen. The results of these measurements are presented in Fig.\ \ref{fig-recPatt}. In Fig.\ \ref{fig-recPatt}(a), the accumulated reception pattern is shown for the \emph{vertical} design, that corresponds to the accumulated signal obtained by adding the individual signals of the eight PDs for a given distance $d$ and orientation $\theta$:
\begin{equation}
S_\mathrm{acc} (d, \theta) = \sum_{i=1}^8 S_i (d, \theta).
\label{Sacc}
\end{equation}
In Figs.\ \ref{fig-recPatt}(b) and \ref{fig-recPatt}(c), an example of the raw signals $S_i$ of the eight PDs, as a function of $\theta$ and for $d = 70$ mm, is shown with the dashed curves with symbols for the \emph{free} and \emph{post} paths, respectively. The corresponding results for the \emph{flower} design are presented in Figs.\ \ref{fig-recPatt}(d), \ref{fig-recPatt}(e) and \ref{fig-recPatt}(f). From Figs.\ \ref{fig-recPatt}(a) and \ref{fig-recPatt}(d) for $S_\mathrm{acc}$, one can clearly distinguish that the accumulated reception pattern for the \emph{vertical} design is more homogeneous with $\theta$ than that from the \emph{flower} design, where one can clearly see a pronounced reception where the PDs are positioned, as they lie horizontally and perpendicular to the mirror. Nonetheless, both designs show omnidirectional reception without blind spots, at least within the measured range. In the next section, we will develop mathematical models for the mean response of a PD in each design. With this, we will compare their performance for distance and orientation detection.

% ======= FIGURE =======
\begin{figure*}[thb]
\includegraphics[width=\textwidth] {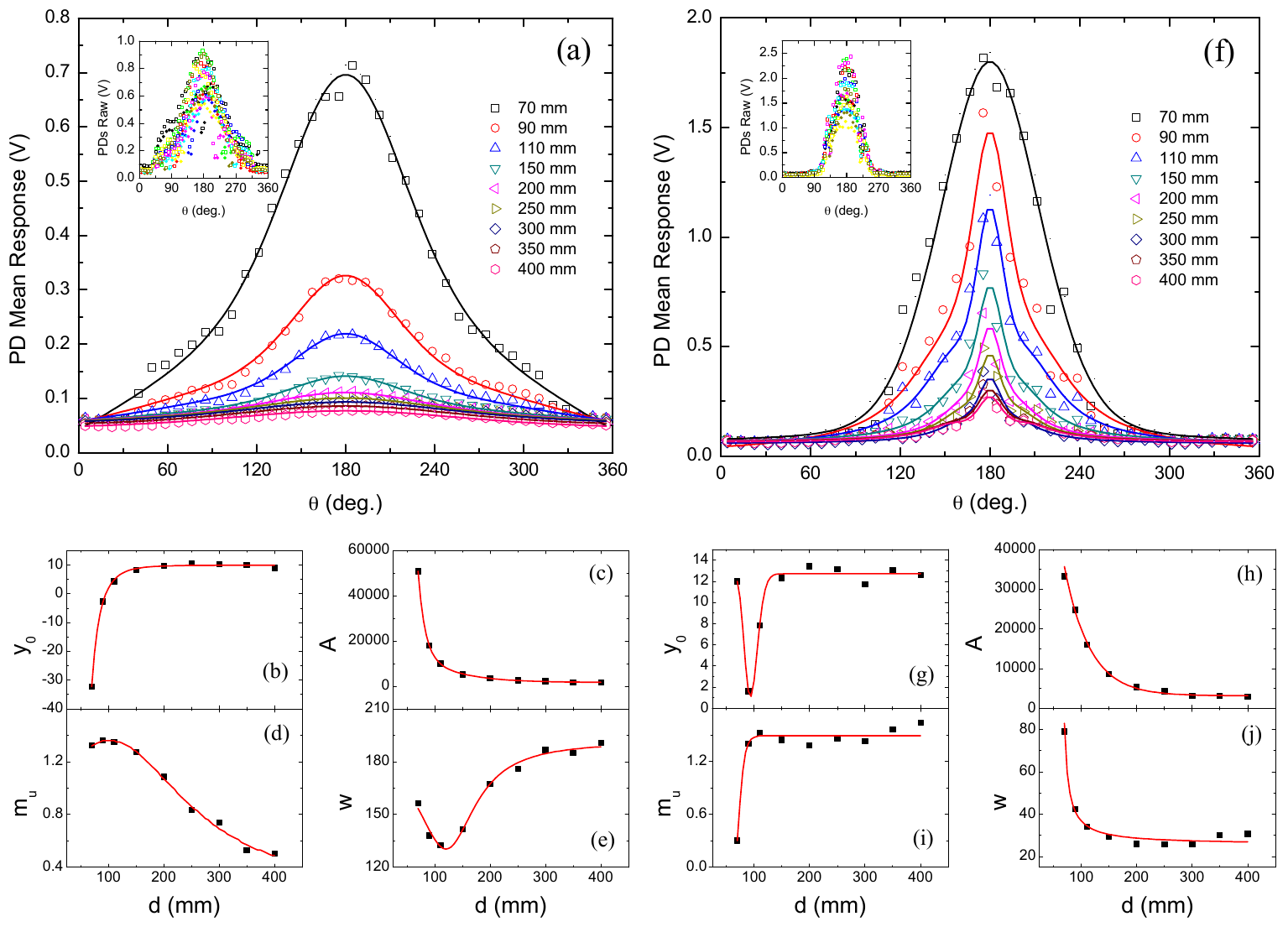}
\caption{In (a) and (f), scattered plots with open symbols for the centered and bin-average smoothed data of the raw signals of the individual PDs, from the \emph{free} and \emph{post} paths combined, as a function of the orientation angle, $\theta$, and for selected distances, $d$, for the \emph{vertical} and \emph{flower} designs, respectively. The solid curves correspond to best fits with the pseudo-Voigt function given in Eq.\ (\ref{psVoigt}). The insets in (a) and (f) show the raw data from a readout of the eight PDs for the \emph{free} path (clear squares) and \emph{post} path (solid circles) after centering their signal peak around 180$^\circ$, but before the smoothing process, for $d=70$ mm. In (b) to (e), nonlinear fits as a function of $d$ of the data for the parameters $y_0$, $A$, $m_u$ and $w$ of the pseudo-Voigt function for the \emph{vertical} design. In (g) to (j), nonlinear fits as a function of $d$ of the data for the parameters $y_0$, $A$, $m_u$ and $w$ of the pseudo-Voigt function for the \emph{flower} design. See text for more details.}
\label{fig-model}
\end{figure*}
% =====================

\section{Model and results}

To compare the performance between the two designs in terms of accuracy for determining distance and orientation, we developed a mathematical model for the mean response of a PD for each design. First, we estimated the offset in orientation for each data set; recall that we have one data set for the \emph{free} path and one for the \emph{post} path for each design. For this, we selected a PD from each data set whose signal peak was closer to $\theta = 180^{\circ}$. Employing the data for all distances for that particular PD, we fitted a Gaussian function to determine the offset angle $\theta_\mathrm{off}$ relative to $180^{\circ}$. This offset angle was then applied to shift $\theta$ for all of the other PDs in the same data set, so that the signal peak of the selected PD was centered around $180^{\circ}$. The peak of the signals from the other PDs are displaced by multiples of $45^{\circ}$ around this PD, consistent with their arrangement in the sensor. Taking this into account, at each distance increment, we proceded to center the signal peak for each individual PD around $180^{\circ}$ by adding or subtracting the appropriate multiple of $45^{\circ}$, depending on its relative position to the selected PD. As an example, for \hbox{$d = 70$ mm}, the \emph{centered} data for the \emph{free} and \emph{post} paths are presented in the insets of Figs.\ \ref{fig-model}(a) and \ref{fig-model}(f) for the \emph{vertical} and \emph{flower} designs, respectively.

Then, in order to smooth the \emph{centered} data, the range $\theta \in [0^{\circ}, 360^{\circ}]$ was divided in 40 bins and the average signal in each bin was calculated, including data from both the \emph{free} and \emph{post} paths. In total, 16 signals were processed for each distance, eight from the \emph{free}-path data and eight from the \emph{post}-path data for each design. Figure \ref{fig-model} shows the results of centering and bin-average smoothing for selected distances in the range $d \in [70 \, \mathrm{mm}, 400 \, \mathrm{mm}]$ with scattered plots marked by open symbols. The function that best models this data is the pseudo-Voigt function \cite{wer74, san97},
\begin{eqnarray}
pV(d, \theta) = & y_0(d) + A(d) \left\{ m_u(d) \frac{2w(d)}{\pi \left[ 4(\theta - \theta_0)^2 + w^2(d) \right]} \right. \nonumber \\
& \left. + [1- m_u(d)] \sqrt{\frac{4 \log 2}{\pi w^2(d)}} e^{-\frac{4 \log 2}{w^2(d)}(\theta - \theta_0)^2} \right\},
\label{psVoigt}
\end{eqnarray}
which is essentially a linear combination of a Lorentzian and a Gaussian functions. In this way, for each fixed distance selected, we fitted the pseudo-Voigt function to the centered and smoothed data as a function $\theta$, with $\theta_0 = 180^{\circ}$, as shown in Figs.\ \ref{fig-model}(a) and \ref{fig-model}(f) with the solid curves for the \emph{vertical} and \emph{flower} designs, respectively. The dependence of the model on the distance, $d$, is therefore encoded in the dependence of the fitting parameters $y_0$, $A$, $m_u$ and $w$ on this variable.

Figures \ref{fig-model}(b) to \ref{fig-model}(e) present nonlinear fits of the parameters $y_0$, $A$, $m_u$ and $w$ of the pseudo-Voigt function, as functions of $d$, for the \emph{vertical} design. The functions used to fit this data are the Logistic function for $y_0$,
\begin{equation}
y_0(d) = \frac{A_1 - A_2}{1 + ( d/d_0)^p + A_2},
\label{Logistic}
\end{equation}
with fitting parameters $A_1$, $A_2$, $d_0$ and $p$. A function with two exponential decay terms for $A$,
\begin{equation}
A(d) = A_0 + A_1 \, e^{-d/t_1} + A_2 \, e^{-d/t_2},
\label{ExpDec2}
\end{equation}
with fitting parameters $A_0$, $A_1$, $t_1$, $A_2$ and $t_2$. The Lorentz function was used for $m_u$,
\begin{equation}
m_u(d) = A_0 + \frac{2 A_1 w_1}{\pi \left[ 4(d - d_0)^2 + w_1^2 \right]},
\label{Lorentz}
\end{equation}
with fitting parameters $A_0$, $A_1$, $w_1$ and $d_0$. The Lorentz function was also used to fit the data for $w(d)$. Meanwhile, Figs.\ \ref{fig-model}(g) to \ref{fig-model}(j) present nonlinear fits of the parameters $y_0$, $A$, $m_u$ and $w$ of the pseudo-Voigt function, as functions of $d$, for the \emph{flower} design. The functions used to fit the corresponding data include the Log-Normal function for $y_0$,
\begin{equation}
y_0(d) = A_0 + \frac{A_1}{\sqrt{2 \pi w_1^2 d^2}} \exp \left[ \frac{-\left( \log \frac{d}{d_0} \right)^2}{2 w_1^2} \right],
\label{LogNormal}
\end{equation}
with fitting parameters $A_0$, $A_1$, $d_0$ and $w_1$. A simple exponential decay was used for $A$,
\begin{equation}
A(d) = A_0 + A_1 \, e^{-d/t_1},
\label{ExpDec1}
\end{equation}
with fitting parameters $A_0$, $A_1$ and $t_1$. The Chapman function was used for $m_u$,
\begin{equation}
m_u(d) = a \left( 1 - e^{-bd} \right)^c,
\label{Chapman}
\end{equation}
with fitting parameters $a$, $b$ and $c$. The Rational function was used for $w$,
\begin{equation}
w(d) = \frac{b + cd}{1 + ad},
\label{Rational}
\end{equation}
with fitting parameters $a$, $b$ and $c$. The resulting model for the mean response of a PD for the \emph{vertical} design is shown in Figs.\ \ref{fig-recPatt}(b) and \ref{fig-recPatt}(c) for $d=70$ mm, where this model is plotted alongside the raw data for each PD, and centered around the corresponding position of each receiver, for the \emph{free} path data in Fig.\ \ref{fig-recPatt}(b) and for the \emph{post} path data in Fig.\ \ref{fig-recPatt}(c). Similarly, for the \emph{flower} design, the model for the mean response of a PD is plotted along the raw data for each PD for the \emph{free} path data in Fig.\ \ref{fig-recPatt}(e), and for the \emph{post} path data in Fig.\ \ref{fig-recPatt}(f), also for $d = 70$ mm.

% ======= FIGURE =======
\begin{figure}[t]
\includegraphics[width=0.5\textwidth] {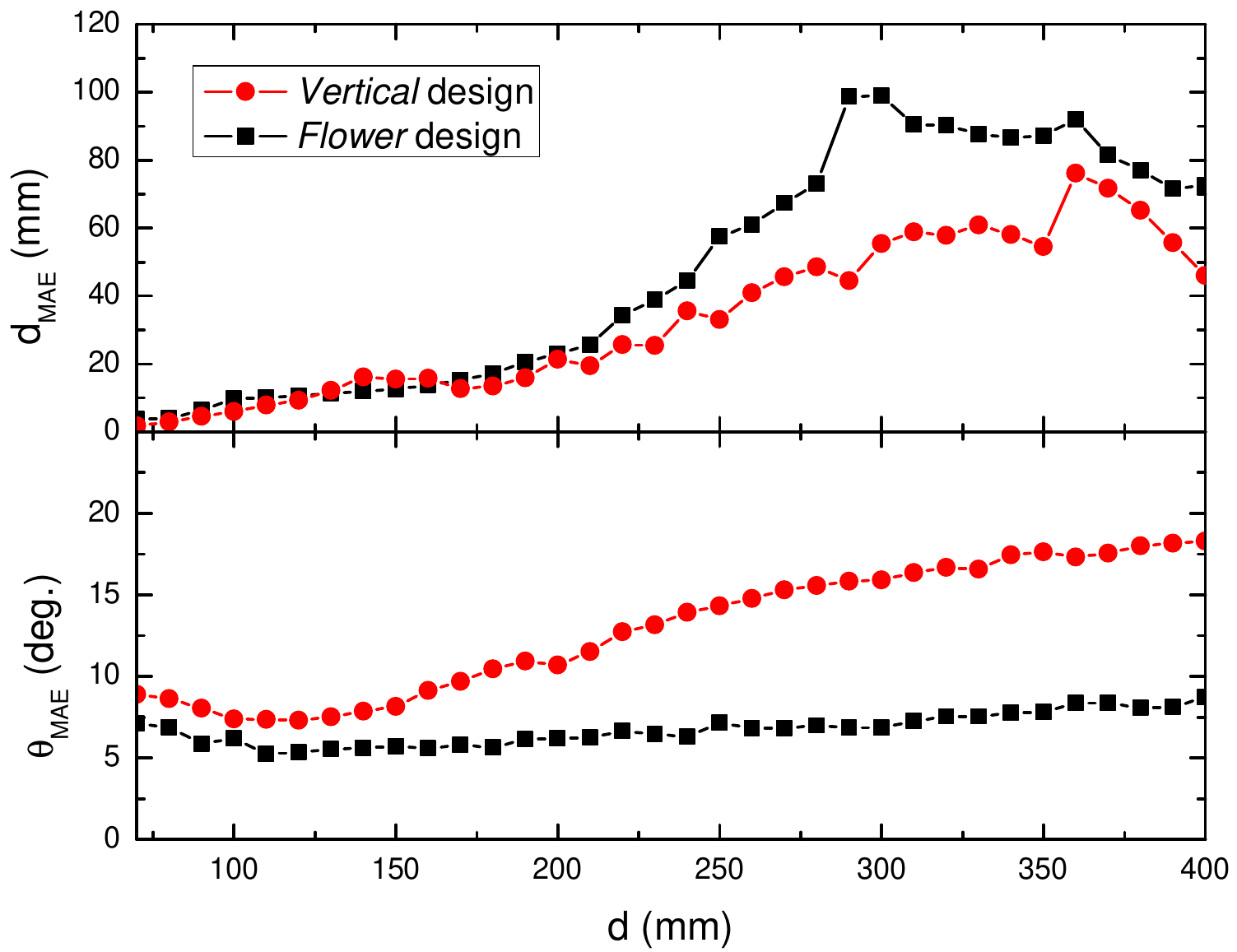}
\caption{Mean absolute error that compares the experimental distance and orientation to those predicted by inverting the model, as a function of the distance $d$ and averaged over the orientation $\theta$ for each distance for the two designs: \emph{vertical} with red solid curves with circles and \emph{flower} with black solid curves with squares.}
\label{fig-MAE}
\end{figure}
% =====================

Armed with these models for the mean response of the PDs for each design, we can now compare the distance $d$ and orientation angle $\theta$ predicted by the respective model against the raw data obtained in a single readout from the eight PDs in our experimental measurements, that is, assessing how well the model can estimate the distance and orientation from raw signals where these quantities are known. It is important to mention that the models are highly nonlinear, making an explicit inversion of them impossible, hence, a numerical analysis is required to obtain this information. Various numerical methods can be considered for this task \cite{numRec}. For example, fitting the raw signal from the eight PDs to the corresponding model, using a \emph{nonlinear least squares in multiple dimensions method} with $d$ and $\theta$ as the fitting parameters. However, this method requires the derivative of the model itself. Conversely, methods that do not require the derivatives, such as the \emph{downhill simplex method in multidimensions}, can be computationally costly; it is necessary to consider that this analysis should be computationally light enough to be implemented on a microcontroller, as those typically employed in the development of mobile-robot prototypes for studies in \emph{swarm robotics}. For this reason, we opted for a simplified approach that minimizes the function
\begin{equation}
\chi_\mathrm{sq} = \sum_{i=1}^{8} \left[ S_i(d, \theta) - pV_i(d, \theta) \right]^2,
\label{minSq}
\end{equation}
where $S_i$ corresponds to the raw signal from each PD, as measured from our experiments, and $pV_i$ corresponds to the relevant model, centered around the position of each PD. We initially minimized this quantity with a \emph{coarse} sweep within the ranges $d \in [70 \mathrm{mm}, 450 \mathrm{mm}]$ and $\theta \in [0^{\circ}, 360^{\circ}]$, with a resolution of 10 mm in distance and approximately 10 mm of arc in orientation, across all of the points from our experimental measurements. Afterwards, we performed a second minimization over all experimental points, employing a \emph{fine} sweep around the minima for $d$ and $\theta$ obtained from the \emph{coarse} sweep, with a range of $\pm 30$ mm in distance and $\pm 30^{\circ}$ in orientation, with a resolution of 2 mm in distance and approximately 2 mm of arc in orientation. Finally, we calculated the mean absolute errors \cite{wil05}
\begin{eqnarray}
d_\mathrm{MAE} = \frac{\sum_{i=1}^{n_{\theta}} | d_\mathrm{exp} - d_\mathrm{mod} |}{n_{\theta}}, \\
\theta_\mathrm{MAE} = \frac{\sum_{i=1}^{n_{\theta}} | \theta_\mathrm{exp} - \theta_\mathrm{mod} |}{n_{\theta}},
\end{eqnarray}
as a function of the distance $d$ by averaging over the orientation $\theta$ for each distance in the experimental data. In the formulas, the subscripts refer to the data from the experimental measurements and the predictions made by the model, while $n_\theta$ corresponds to the number of points with different orientation for a given distance. It should be noted that the mean absolute error was calculated across all of the experimental data measured for a given design, whether \emph{vertical} or \emph{flower}, including both data sets for the \emph{free} and \emph{post} paths. The results are presented in Fig.\ \ref{fig-MAE}, where it can be seen that the \emph{vertical} design resolves the distance better than the \emph{flower} design, while the \emph{flower} design exhibits better resolution in orientation compared with the \emph{vertical} design. Overall, the accuracy of both designs decreases with the increasing distance, however, from the results of Fig.\ \ref{fig-MAE}, it is possible to estimate maximum errors of $\pm 2 \, \mathrm{cm}$ in distance and $\pm 10^{\circ}$ in orientation for $d < 20$ cm, and maximum errors of $\pm 10 \, \mathrm{cm}$ in distance and $\pm 18^{\circ}$ in orientation for $d \in [20 \, \mathrm{cm}, 30 \, \mathrm{cm}]$ for both designs. We consider these results to be good, taking into account that other designs that employ discrete sensors, as those discussed before, compensate their blind spots and noise in the measurements with multiple readouts, complimented with some scanning motion. A direct comparison with the resolution of other vision sensors was not possible as, for the platforms in Table \ref{tab-plats}, only Rice r-one reports a resolution in orientation of 22.5$^{\circ}$; this resolution is worse than the resolution of our sensors at the same distance. Notice that the resolution in orientation for our \emph{flower} design is basically independent on the distance, and from Fig.\ \ref{fig-MAE}, has a maximum error of about $\pm 8^{\circ}$ for all distances.

On another point, we are aware that the error analysis we developed here is coarse, nonetheless, it is sufficient to be able to compare the detection capacity between the two designs introduced in this work, as well as to point out different directions to improve on them. Although a more detailed error analysis, with error propagation would be desirable, all the uncertainties in our measurements correspond to systematic errors, encompassed in the \emph{mean absolute error} analysis we developed.

\section{Conclusions}

In this work, we introduced and studied two designs for camera-less omnidirectional catadioptric vision sensors made with discrete components: one infrared LED and eight infrared photodiodes, and one custom-made mirror with rotational symmetry. We developed the methods necessary to design and fabricate the mirrors, allowing for any interaction range to be covered, limited only by the sensitivity and radiant intensity of the PDs and LED, respectively. The vision sensors developed are suitable for autonomous mobile-robot prototypes as those typically used in studies of \emph{swarm robotics} on a plane, and could be easily employed for communication purposes if a suitable protocol is implemented in the detection method. Our results demonstrate that the \emph{vertical} design offers better resolution in determining the distance from a single readout of the photodiodes, while the \emph{flower} design is more effective for determining orientation among robotic agents. Improvements in accuracy could be achieved by implementing techniques previously used by other platforms, where the robots stop and perform a measurement sweep to reduce noise in the signals received. Additionally, the development of a hybrid sensor, with photodiodes arranged in both vertical and horizontal orientations may be worth some consideration.

As a final remark, we would like to mention that these types of sensors, akin to biologically inspired vision sensors with a discrete number of receivers, are popular and essential in the development of robotic platforms for studies in swarm robotics. Given our results, we believe that our designs and methods may prove useful for future applications in this field.

\ack 
The authors acknowledge financial support from the grant VIEP-BUAP 2023-00310. VD acknowledges support from the ICTP through the Associates Programme (2022-2027), as well as financial support from CONACyT through the grant 257352. JFC-M acknowledges support from CONACyT through the scholarship 811750. VD thanks \hbox{M.\ E.\ Delmas-Pfingsten} for proofreading the manuscript.

\end{document}